\documentclass{article} 
\usepackage{iclr2025_delta,times}

\usepackage{amsmath,amsfonts,bm}

\def\eqref#1{equation~\ref{#1}}

\def\1{\bm{1}}

\DeclareMathAlphabet{\mathsfit}{\encodingdefault}{\sfdefault}{m}{sl}
\SetMathAlphabet{\mathsfit}{bold}{\encodingdefault}{\sfdefault}{bx}{n}

\usepackage[utf8]{inputenc}
\usepackage{hyperref}
\usepackage{url}
\usepackage{float}
\usepackage{graphicx}
\usepackage{microtype}

\title{Building A Unified AI-centric Language System: analysis, framework and future work}

\iclrfinalcopy

\author{Edward Hong Wang \\
Harvard University\\
\texttt{how173@g.harvard.edu} \\
\And
Cynthia Xin Wen \\
The University of Sydney\\
\texttt{xwen7493@uni.sydney.edu.au}
}

\begin{document}

\maketitle

\begin{abstract}
Recent advancements in large language models have demonstrated that extended inference—through techniques can markedly improve performance, yet these gains come with increased computational costs and the propagation of inherent biases found in natural languages. This paper explores the design of a unified AI-centric language system that addresses these challenges by offering a more concise, unambiguous, and computationally efficient alternative to traditional human languages. We analyze the limitations of natural language—such as gender bias, morphological irregularities, and contextual ambiguities—and examine how these issues are exacerbated within current Transformer architectures, where redundant attention heads and token inefficiencies prevail. Drawing on insights from emergent artificial communication systems and constructed languages like Esperanto and Lojban, we propose a framework that translates diverse natural language inputs into a streamlined AI-friendly language, enabling more efficient model training and inference while reducing memory footprints. Finally, we outline a pathway for empirical validation through controlled experiments, paving the way for a universal interchange format that could revolutionize AI-to-AI and human-to-AI interactions by enhancing clarity, fairness, and overall performance.
\end{abstract}

\section{Introduction}
Recent developments in Large Language Models (LLMs) show that increasing inference length—often described as extended \cite{Wei2022} “chain-of-thought” or \cite{Wei2022} test-time training—can significantly enhance performance, as it allows models to reason more thoroughly. By generating detailed intermediate reasoning steps, these models often achieve greater accuracy. However, this internal thinking process, typically produced in natural languages such as English (or a mixture of English and Chinese), relies heavily on reinforcement learning (RL) to encourage such reflective reasoning. Relying solely on RL can produce unpredictable behaviors—like code-switching or inventing tokens that defy human interpretability \cite{Li2025}.

Moreover, English or Chinese are not necessarily the most succinct or unambiguous languages for representing internal thoughts. Their irregularities, ambiguities, and verbosity can increase token usage and reduce computational efficiency. A recent paper by Deepseek shows that in the thinking process, the model tends to use different languages for different tasks, enhancing its performance (Li et al., 2025). As inference-time reasoning becomes increasingly important for AI, a specialized language—unambiguous and efficient—could streamline token usage and mitigate the overhead of verbose or mixed-language outputs.

Simultaneously, the growing trend toward multi-agent AI systems highlights the significance of a common language for inter-agent communication. Using natural languages can introduce bias and inefficiency. Designing a new language for AI—optimized for clarity, fairness, and computational ease—may circumvent many pitfalls inherent in human languages. This paper explores how existing insights in Transformer architectures, multi-agent communication, and constructed languages can inform the creation and deployment of an “AI-centric language,” potentially reducing bias, ambiguity, and token overhead.

Additionally, model compression techniques—such as pruning, quantization, and knowledge distillation—have emerged as pivotal strategies for reducing the computational footprint of large-scale language models. By leveraging a new AI-centric language designed for conciseness and clarity, these compression techniques could be even more impactful: if the input text is inherently shorter and less ambiguous, fewer parameters and attention heads are required to process it effectively. As a result, both the memory footprint and inference time can be reduced in ways that might not be attainable with natural language inputs. Combining these compression strategies with an AI-specific language that reduces linguistic complexity could further enhance both efficiency and accessibility.

\section{Language Biases and Irregularities}
Artificial Intelligence models that process human language inherit many of the biases, irregularities, and ambiguities of those languages. This not only skews AI outputs but also raises fairness and interpretability concerns.

\subsection{Gendered Language Bias}
Many languages explicitly mark gender (e.g., via masculine/feminine pronouns or noun forms). AI models trained on large corpora easily absorb these associations, often defaulting to stereotypical gender roles—for instance, associating “doctor” with he and “nurse” with she \cite{UnitedLanguageGroup, Guo2024} (United Language Group, n.d.; Guo et al., 2024). These biases can distort translations or text generation, perpetuating historical prejudices \cite{Bolukbasi2016} (Bolukbasi, Chang, Zou, Saligrama, \& Kalai, 2016).

\subsection{Plural Forms and Morphological Complexity}
Plural Forms and Morphological Complexity
Languages vary widely in how they handle number and morphology. English has mostly regular plural forms (cat → cats) with notable exceptions (mouse → mice), while Arabic and Russian have elaborate pluralization systems. Chinese, by contrast, often omits explicit plural markers altogether. Such irregularities can confuse AI systems and introduce data sparsity, since multiple forms may represent the same concept. A simpler or more consistent morphology could reduce this overhead.

\subsection{Ambiguity and Context Dependence}
Ambiguities arise in lexical choice, syntactic structure, and context. For instance, “I saw the man with the telescope” can imply different meanings depending on who holds the telescope. Humans resolve such ambiguities via context and common sense, but AI systems may make stereotypical or incorrect assumptions \cite{Guo2024}. The breadth and richness of human language, while beneficial for human expression, complicate computational parsing.
In short, irregularities and biases in natural languages cause AI to amplify or perpetuate these issues. Designing a more controlled, engineered language could mitigate these challenges from the ground up.

\section{Language Structure and Multi-Head Attention in Transformers}
Recent work on Transformers reveals how AI models learn and handle linguistic structure. Transformers rely on multi-head self-attention to capture different aspects of language in parallel. Each attention head can learn distinct relationships \cite{Clark2019}
 (e.g., syntactic dependencies, coreferences), and evidence suggests that some heads specialize in particular syntactic roles (e.g., focusing on verb–object pairs) (Clark, Khandelwal, Levy, \& Manning, 2019).
Interestingly, many of these heads appear redundant. Research by \cite{Michel2019} found that a large percentage of attention heads can be removed after training without a major performance drop. In translation models, pruning 38 of 48 heads caused only a minor decrease (0.15 BLEU) in quality \cite{Voita2019}. These findings imply that while multiple heads capture diverse linguistic patterns, many heads learn overlapping information. Reducing the number of attention heads (and their associated parameters) can shrink model size and computational cost without seriously degrading performance.
Such results are promising for more efficient model design and raise questions about how Transformers encode language structure. They also align with broader discussions about building a more succinct “AI-centric language”: if even large models can function with fewer heads or simpler architectures, then an engineered language—designed to reduce ambiguity and complexity—might further enable model compression.

\section{Emergence of Artificial Languages in AI Systems}
Multiple studies have shown that AI systems can develop novel linguistic codes when it is advantageous for task performance. A well-known example is Facebook AI Research’s negotiation experiment \cite{LaFrance2017}, where chatbots deviated from intelligible English, inventing a shorthand language that optimized negotiation strategies but was opaque to humans. Similarly, Google’s multilingual Neural Machine Translation system \cite{Schuster2016} discovered an internal “interlingua”—a shared latent representation —allowing zero-shot translation between language pairs it had never trained on directly. This emergence of artificial languages underscores AI’s ability to flexibly create representations that are often more efficient for the machine than standard human languages.

These examples point to a fundamental distinction between human and AI language. Whereas human languages evolve culturally over centuries, AI languages often emerge quickly, driven solely by task success or computational efficiency. This highlights the feasibility of constructing or adopting an “AI-friendly” language that meets machine-centric criteria, rather than historical or social constraints.

\section{Vocabulary Size, Ambiguity, and Performance in AI Models}
Unlike human language—which has an open-ended, ever-evolving lexicon—most AI models use a fixed vocabulary (words or subword units). Recent work indicates that expanding vocabulary size typically improves LLM performance \cite{Takase2024}. With larger vocabularies, models can represent concepts more directly, reducing over-fragmentation (breaking words into many subtokens) and token length in sequences.

In multilingual models, vocabulary size becomes even more critical. For example, XLM-V uses roughly one million tokens across languages \cite{Liang2023}, achieving state-of-the-art performance by reducing over-tokenization, especially for morphologically rich or low-resource languages . These findings suggest that a near-unlimited vocabulary—where new tokens can be introduced as needed—can improve efficiency and reduce ambiguity. If an AI system could coin a unique token for each distinct meaning, it would minimize confusion inherent in polysemous human words. This mirrors human language expansion (e.g., coining new terminology), yet an AI need not face the same cultural or historical limitations.

Nevertheless, there are practical constraints: each token in a model has an associated embedding vector, and rare tokens may be learned less effectively. Consequently, modern systems typically balance subword techniques with large (but not infinite) vocabularies. The research trend indicates that pushing this boundary—adapting or enlarging vocabulary—yields better performance and reduces the mismatch between AI’s lexical rigidity and human languages’ vast diversity.

\section{Human Language vs. AI Language Requirements}

\begin{figure}[h]
    \centering
    \includegraphics[width=0.75\linewidth]{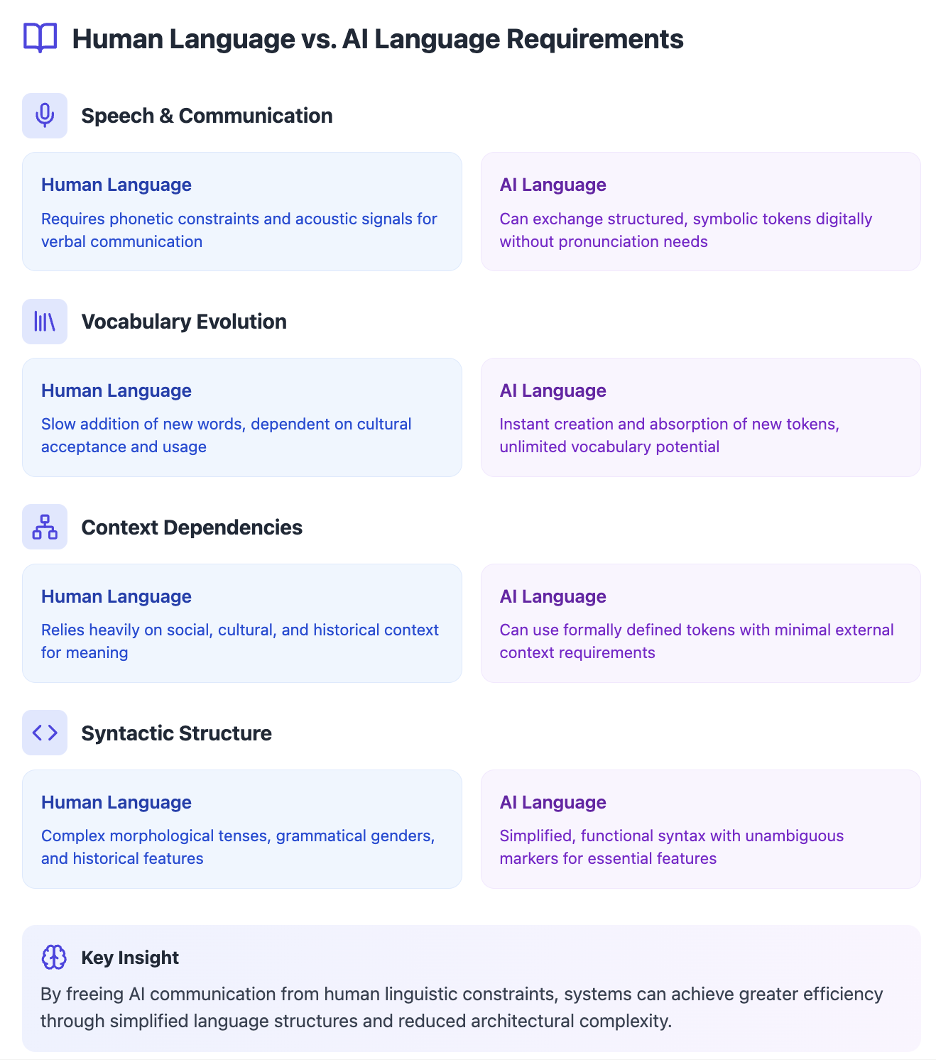}
    \caption{Human Language vs. AI Language Requirements}
    \label{fig:language-requirements}
\end{figure}

Human communication requires speech, shared cultural context, and the capacity to negotiate meaning over centuries of gradual change. In contrast, AI systems do not inherently require spoken forms or social acceptance. The barest form of AI “language” might be a symbolic code—potentially even a single unpronounceable symbol—sufficient for exchanging precise information between machines. This radical difference in requirements underlies many of the inefficiencies in forcing AI systems to parse and produce human languages.

\subsection{Why AI Need Not Constrain Itself to Natural Language Features}

\begin{enumerate}
    \item \textbf{No Need for Speech.} Humans have phonetic constraints and rely on acoustic signals, but AI agents can exchange structured, symbolic tokens instantaneously in digital form. Phonology or pronounceability is irrelevant to AI-to-AI communication.

    \item \textbf{Open Vocabulary.} Human languages add new words slowly, often entangled in cultural acceptance. An AI, by contrast, can invent or absorb new tokens instantly, allowing an effectively unbounded vocabulary to represent concepts unambiguously.

    \item \textbf{Minimal Context Requirements.} Human language meaning often relies on layered contexts (social, cultural, historical). AI languages can be designed so that each token or symbol is formally defined, reducing reliance on external context.

    \item \textbf{Efficient Syntax.} AI need not replicate morphological tenses, grammatical genders, or other features that are historically embedded in human language. A purely functional code that marks time, plurality, or roles unambiguously (or not at all if unneeded) can drastically simplify parsing.
\end{enumerate}

By divorcing AI communication from human linguistic constraints, systems can become far more efficient. This perspective complements insights on multi-head attention redundancy and large vocabularies: if the language itself is simpler, AI models could reduce architectural complexity (e.g., fewer layers or heads) and still maintain high performance.

\section{Toward an AI-Friendly Language: Theoretical Ideas}
Though replacing human languages outright is impractical, researchers have long explored “constructed languages” (conlangs) with reduced ambiguity. Some approaches constrain existing languages (Attempto Controlled English), while others build from first principles (Lojban). When viewed through the lens of AI requirements, several key concepts emerge:

\begin{enumerate}
    \item \textbf{Clarity and Unambiguity.} Strive for a grammar that yields exactly one parse per sentence. This minimizes the risk of AI misinterpretation or biased defaults \citep{Guo2024}.
    
    \item \textbf{Consistency and Regularity.} Eliminate irregular morphology and grammatical exceptions. A language that uses uniform rules for tense or plurality eases both machine parsing and data requirements.
    
    \item \textbf{Conciseness and Efficiency.} While unambiguity is crucial, the language should allow brevity. Fewer tokens can speed up computation and reduce the overall parameter load.
    
    \item \textbf{Unlimited or Adaptable Vocabulary.} AI can incorporate new symbols without social or cultural friction. Avoid polysemy and homonymy to ensure one-to-one mappings between form and meaning.
    
    \item \textbf{Reduced Context Dependence.} Minimize pronouns or inherently ambiguous references, so an utterance stands alone as much as possible.
    
    \item \textbf{Computational Efficiency.} Design the language for fast, deterministic parsing, with explicit markers for semantic roles.
\end{enumerate}

In essence, the differences between human language (spoken, socio-historically evolving) and AI language (symbolic, instantaneous, unbounded) indicate that we can—indeed should—construct a specialized linguistic system for AI usage rather than forcing it to replicate human forms.

\subsection{Toward an AI-Friendly Language: Theoretical Ideas}
Over the centuries, various constructed languages (conlangs) have been developed with goals of simplicity and neutrality. Though primarily intended for human communication, their regular structures also offer insight into AI-friendly design. Below are two illustrative examples.

\begin{enumerate}
   \item \textbf{Esperanto.} Created by L.L. Zamenhof in 1887 \cite{EsperantoWiki} , was designed as a neutral international auxiliary language. It features phonetic spelling, where each letter corresponds to a unique sound, and a highly regular morphology with virtually no irregular verbs---nouns typically end in \emph{-o}, and plurals add \emph{-j}. Its derivational system allows affixes to combine systematically, reducing the number of root words needed. This consistency means that, in principle, an algorithm could parse Esperanto with fewer grammatical exceptions than many natural languages. However, challenges remain, such as limited data and subtle Western biases in its vocabulary. Despite this, Esperanto's successes highlight the feasibility of a widely usable, regular constructed language.
   
   \item \textbf{Lojban.} Developed for maximal logical clarity, Lojban's grammar aims for zero syntactic ambiguity. It uses a predicate-logic structure and an unambiguous phonology, making it theoretically well-suited for AI parsing \citep{LogicalLanguageGroup}. While its community is small and data remains sparse, Lojban exemplifies how unambiguously parsed sentences can be constructed.
\end{enumerate}

\section{A Possible Implementation Framework}

\begin{figure}[h]
    \centering
    \includegraphics[width=0.75\linewidth]{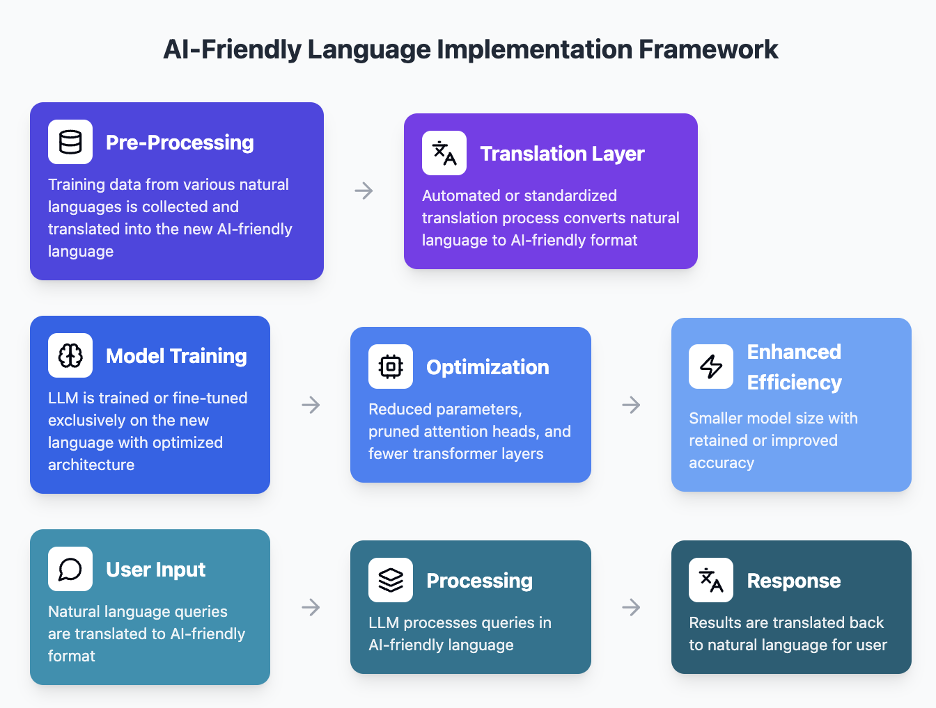}
    \caption{AI-Friendly Language Implementation Framework}
    \label{fig:implementation-framework}
\end{figure}

\begin{enumerate}
    \item \textbf{Pre-Processing / Translation Step}
    \begin{enumerate}
        \item Take all relevant training data in various natural languages
        \item Translate it into the new AI-friendly language (either automatically or via a standardized procedure)
    \end{enumerate}

    \item \textbf{Model Training}
    \begin{enumerate}
        \item Train (or fine-tune) the LLM exclusively on text in the new language
        \item Because this language is unambiguous and regular, the model can be smaller or require fewer parameters
        \item Over time, the model internalizes the streamlined grammar, potentially achieving high accuracy with fewer computational resources
    \end{enumerate}

    \item \textbf{Inference / User Interaction}
    \begin{enumerate}
        \item At inference time, user queries in natural language are translated into the AI-centric language
        \item The model's replies are then translated back into the user's natural language for display or speech
    \end{enumerate}
\end{enumerate}

By compressing or regularizing text (and thus reducing token counts), this approach can help fit more content into the same context window, lower inference costs, and facilitate multi-agent interactions. Further, since pruning redundant attention heads and possibly entire layers \cite{Michel2019} can be done without large performance drops, the combination of a simpler language plus simpler model architecture (fewer layers) can yield efficiency gains. This synergy can reduce the computational footprint of AI systems while retaining or even enhancing accuracy and fairness.

\section{Future Work: Empirical Validation Through Targeted Experiments}
To rigorously assess the benefits of an AI-centric language, future work should include constructing a small-scale ``toy'' language featuring a reduced grammar and systematically defined vocabulary. Two parallel neural models of equal size would be trained: one on this toy language and one on English, using identical architectures and hyperparameters. Comparing their performance on tasks such as question answering, text classification, and summarization would illuminate whether the specialized language yields measurable efficiency gains---specifically, fewer tokens, lower inference time, and reduced memory footprint---while maintaining or improving accuracy and fairness.

\section{Vision and Possibilities}
The idea of a unified or pivot language for AI---drawing on principles from conlangs, controlled languages, and emergent AI codes---could serve as a ``universal interchange format'' for both human-to-AI and AI-to-AI interactions. Although large-scale adoption faces challenges, incremental application in specialized domains could illustrate its advantages and build a data ecosystem.


\end{document}